\documentclass{article}

\usepackage[numbers]{natbib}
\PassOptionsToPackage{numbers, compress}{natbib}
\usepackage[preprint]{nips_2018}

\usepackage[utf8]{inputenc} 
\usepackage[T1]{fontenc}    
\usepackage{hyperref}       
\usepackage{url}            
\usepackage{booktabs}       
\usepackage{amsfonts}       
\usepackage{nicefrac}       
\usepackage{microtype}      
\usepackage{amsmath}
\usepackage{amssymb}
\usepackage{booktabs}
\usepackage{multirow}
\usepackage{todonotes}

\title{Adversarial attacks against\\Fact Extraction and VERification}

\author{
  James Thorne\\
  University of Cambridge \\
  \texttt{jt719@cam.ac.uk} \\
  \And
  Andreas Vlachos \\
  University of Cambridge \\
  \texttt{av308@cam.ac.uk}
}

\begin{document}




\maketitle

\begin{abstract}
  This paper describes a baseline for the second iteration of the Fact Extraction and VERification shared task (FEVER2.0) which explores the resilience of systems through adversarial evaluation. We present a collection of simple adversarial attacks against systems that participated in the first FEVER shared task. FEVER modeled the assessment of truthfulness of written claims as a joint information retrieval and natural language inference task using evidence from Wikipedia. 
  A large number of participants made use of deep neural networks in their submissions to the shared task. 
  The extent as to whether such models understand language has been the subject of a number of recent investigations and discussion in literature. 
  In this paper, we present a simple method of generating entailment-preserving and entailment-altering perturbations of instances by common patterns within the training data. We find that a number of systems are greatly affected with absolute losses in classification accuracy of up to $29\%$ on the newly perturbed instances. Using these newly generated instances, we construct 
  a sample submission for the FEVER2.0 shared task. Addressing these types of attacks will aid in building more robust fact-checking models, as well as suggest directions to expand the datasets.
\end{abstract}

\section{Introduction}
Significant progress for a large number of natural language processing tasks has been made through the development of new deep neural models; higher scores for shared tasks such as Natural Language Inference \cite{Bowman2015} and Question Answering
\cite{Rajpurkar2016} have been achieved through models which are becoming increasingly complex. This complexity raises new challenges: as models become more complex, it becomes difficult to fully understand and characterize their behaviour. This has implications from a statistical learning perspective as the more complex models can be prone to brittleness. From an NLP perspective, there has been an ongoing discussion as to what extent these models understand language \cite{Jia2017} or to what extent they are exploiting unintentional biases and cues that are present in the datasets they are trained on \cite{Poliak2018, Gururangan2018}. When the model is evaluated on data outside of the distribution defined (implicitly) by its training dataset, 
its behaviour is likely to be unpredictable; such ``blind spots'' can be exposed through \emph{adversarial evaluation} \cite{szegedy2014}.

In this paper we focus on the use of adversarial evaluation in the context of fact checking: this is the task of predicting whether a claim is supported or refuted 
by evidence (see \cite{Thorne2018} for an in-depth task description). As automated systems for fact checking have potentially sensitive applications 
it is important to study the vulnerabilities of these systems, as well as the deficiencies of the datasets they are trained on. 
Through adversarial evaluation, where systems are exposed to examples of statements intentionally created to result in incorrect predictions, we can better understand the limitations of the systems and possibly use these adversarial instances to regularize the model through training data augmentation \cite{szegedy2014, goodfellow2015, Miyato2017}.   

The Fact Extraction and VERification (FEVER) shared task \cite{Thorne2018b} invited participants to build systems to predict whether short factoid sentences (referred to as claims) -- manually written using information derived from Wikipedia -- 
are \textsc{Supported} or \textsc{Refuted} by evidence or whether there is \textsc{NotEnoughInfo} to reach a conclusion. Most participants in the shared task modeled it as pipeline consisting of an information retrieval component -- finding pages/sentences with content relevant to a given claim -- and a Natural Language Inference (NLI) component -- proving that given the evidence, a claim is either \textsc{Supported} or \textsc{Refuted}. The second iteration of the task (FEVER2.0) is building on the same dataset, but takes the form of a build-it, break-it, fix-it task \cite{ettinger2017toward}. \emph{Builders} will create 
systems based on the original FEVER dataset and task definition; \emph{breakers} will generate adversarial examples targeting the systems built in the first stage; finally, \emph{fixers} will implement solutions to remedy the attacks from the second stage.

\begin{figure}[]
\centering
\bgroup
\def\arraystretch{1.5}
\begin{tabular}{|p{6cm}|p{6cm}|}
\hline
\multicolumn{2}{|l|}{\textbf{FEVER Claim:} Bullitt is a movie directed by Phillip D'Antoni} \\
\multicolumn{2}{|p{12cm}|}{\textbf{Evidence:} Bullitt is a 1968 American action thriller film directed by Peter Yates and produced by Philip D'Antoni} \\
\multicolumn{2}{|l|}{\textbf{Label:} \textsc{Refuted}} \\ \hline
\textbf{Entailment Preserving Transformation} & \textbf{Entailment Altering Transformation} \\
There is a movie directed by Phillip D'Antoni, it is called Bullitt. & Bullitt is not a movie directed by Phillip D'Antoni \\
\textbf{New Label:} \textsc{Refuted} & \textbf{New Label:} \textsc{Supported} \\ \hline
\end{tabular}
\egroup
\caption{Adversarial instances generated through rule-based transformations of existing claims}
\label{fig:example_attack}
\end{figure}

We describe a baseline for the \emph{breaking} stage FEVER2.0 shared task: we use simple, rule-based transformations of claims from the original FEVER dataset to generate new ones from existing instances in the dataset. This approach makes use of the same evidence, avoiding the need for additional evidence finding, and the rules can either change or preserve their label with respect to the evidence (see Figure~\ref{fig:example_attack}). 
We also introduce metrics for scoring the \emph{potency} of the attacks (which accounts for the number of systems that incorrectly classify the newly generated adversarial instances). We attack a number of state-of-the-art models: the Neural Semantic Matching Network \cite{nie2019combining}, HexaF Model \cite{yoneda-EtAl:2018:FEVER}, Athene \cite{hanselowski-EtAl:2018:FEVER} and Papelo \cite{malon:2018:FEVER} -- the highest ranked systems for the first FEVER shared task -- as well as an ESIM model \cite{Chen2016} with ELMo embeddings \cite{Peters2018a} and a Decomposable Attention model \cite{Parikh2016} -- built on the baseline architecture presented in \cite{Thorne2018a}. We evaluated the effect of the rule-based transformations 
on both the Information Retrieval and NLI components of these systems and found that all systems achieved lower accuracy
on the new data we generated with decreases of between $11.32\%$ and $29.16\%$. Also, while the information 
retrieval components of some systems were not affected to the same extent, our experimental results indicate that some implementations were more sensitive the new, adversarial, data than others. 
We hope that this baseline implementation and report serves as a useful resource for future participants of the FEVER2.0 shared task by showing that a relatively simple method reasonably effective and highlighting in what ways the models we test against are affected. To help participants, we release dockerized versions of the systems we test against \footnote{\url{https://github.com/j6mes/fever-docker}} as well as code for generating and evaluating adversarial examples \footnote{\url{https://github.com/j6mes/fever2-baseline}}.


\section{Related Works}
Adversarial examples have been initially 
studied in the field of computer vision.
\citet{szegedy2014} identified an over-sensitivity of state-of-the-art models whereby perturbations to the image (by altering pixel intensities in a way which is imperceptible to humans) resulted in changes to the predictions made by the model. These perturbation were generated through altering the model input to maximize the classification error. 
While the proposed methods of altering pixel intensities was generally imperceptible to humans, making similar perturbations to text is more challenging due to the discrete symbol space and the need to preserve grammaticality: modifying a single token may either change the label of the instance or introduce grammatical errors. Various methods for successful attacks have been proposed, including manual construction, character perturbations, addition of distractor information, rule-based transformations and paraphrasing, and automated generation. In our brief survey, we compare these methods and the trade-off between the level of automation (which allows both scale and diversity of new claims) and whether the perturbation unintentionally changes the label of the instance or induces grammatical error, which would require human annotation to identify and resolve.  

\paragraph{Manual Construction}
Small adversarial datasets have been manually constructed and successfully used to identify limitations in Machine Translation  \cite{Burlot2017,isabelle2017challenge}, Sentiment Analysis \cite{mahler2017breaking, staliunaite2017breaking} and Natural Language Understanding \cite{Levesque2013} systems. Instances are generated that exploit world knowledge, semantics, pragmatics, morphology and syntactic variations. By manually constructing adversarial instances, the attacker would have a high degree of confidence that the text will be grammatical and that instances will be correctly labeled. However, exploiting human knowledge of language is comparatively expensive and is difficult to scale to construct larger datasets. 

\paragraph{Character-level Perturbation}
Character-level attacks have also highlighted the brittleness of NLP systems: by making letter swaps or insertions, \citet{Belinkov2018} and \citet{Ebrahimi2018} have generated distorted examples which cause misclassifications or translation errors. While it is unlikely that a single character can unintentionally change the meaning of a sentence, this method is still \emph{intentionally} introducing errors. 

\paragraph{Distractor Information}
\citet{Jia2017} evaluated the addition of distractor information in reading comprehension systems. Adversarial instances are generated for the SQuAD \citep{Rajpurkar2016} shared task (question answering against a short passage of text) 
by concatenating short distractor sentences to the passage. The distractor sentences are generated through perturbing the question with entity substitution and generating a false 
answer which has a similar form to the actual answer with rule-based substitutions. 
Furthermore, the additional 
information concatenated to the original passage of text is by construction irrelevant, as it is about another entity. Thus it is unlikely to cause a change to the meaning of the text requiring the instance to be labelled. Additionally, this approach does not require manual generation of the instances (human annotators are only used for filtering out ungrammatical distractors) this would be less expensive than manual construction of adversarial instances meaning that greater scale can be achieved without the risk of annotators unintentionally introducing a bias. 

\paragraph{Paraphrasing}
\citet{iyyer2018adversarial} and \citet{Ribeiro2018adebug} apply paraphrase-based transformations to generate adversarial instances using alignments from parallel corpora from translation tasks. \citet{iyyer2018adversarial} attack sentiment analysis 
and recognizing textual entailment systems, 
generating paraphrases of instances with a encoder-decoder model architecture. 
In the process of generating adversarial instances, the meaning could be altered requiring relabelling (for example a sentence pair with an \textsc{Entailment} relation may become \textsc{Neutral} after perturbation) or the newly generated text could be ungrammatical. In an error analysis, the authors identified that in 17.7-22.3\% of cases, the generated examples were not paraphrases and in 14.0-19.3\% of cases, the paraphrases were ungrammatical. \citet{Ribeiro2018adebug} evaluate phrase substitutions, obtained from alignments from a translation task, as a method of generating semantically equivalent instances for visual QA, sentiment analysis, and reading comprehension. The authors incorporate additional filtering to remove instances that are ungrammatical or unnatural: rather than using human annotators, this is automated through computing the probability of translating the original instance and getting the paraphrase when back-translating. 

\paragraph{Programmatic Construction of Adversarial Dataset}
\citet{Naik2018} introduce a stress test evaluation dataset for NLI containing a number of methods for generating new claims that exploit limitations and biases present in state-of-the-art models in the context of the MultiNLI \cite{williams2017broad} shared task. The adversarial dataset was constructed by applying three types of transformations to the MultiNLI development data split: meaning-altering transformations are performed to instances that require numerical reasoning through rule-based transformation; distractor phrases that preserve meaning are appended to instances (that exploit models' biases for strong indicators for negation and sentence length); finally, perturbations to some instances are introduced to mimic typographical errors. While the rule-based changes would preserve the label and are grammatical, some of the changes in this dataset are not natural (for example, the word overlap rule which appended `and true is true' tautology is logically correct but unlikely to occur in everyday language). This approach is relatively cheap in comparison to manual construction as one rule can be applied to many instances. 

\paragraph{Automated Generation}
\citet{Zhao2018} generate natural language adversaries for an NLI task through the use of an autoencoder architecture. The new instances exhibit high diversity and are regarded as ``natural'' with 86\% of human annotators stating that the new instances were grammatical and 81\% stating that the new instances was similar to the original on a pilot study of 20 examples. However, while this method will generate instances that are similar to the original, it is not certain that the label for the newly generated instances will be preserved as similarity does not guarantee semantic equivalence. 
\citet{Zellers2018} introduce an adversarial method for generating negative examples for a multiple-choice answer selection task through generating probable sentences (with the aid of a language model) that induce misclassifications. 
Using this approach, however, it is not possible to generate examples that preserve or deterministically alter the entailment relation: again requiring the newly generated instances to be labeled by humans.



\section{Adversarial Attacks Against FEVER \\(The \emph{Break-it} Phase of the FEVER2.0 Shared Task)}

\begin{figure}
    
\bgroup
\def\arraystretch{1.5}
    \begin{tabular}{|p{6cm}|p{6cm}|}
    \hline
    \multicolumn{2}{|l|}{\textbf{FEVER Claim:} Lily James has been on TV.} \\
    \multicolumn{2}{|l|}{\textbf{Label:} \textsc{Supported}} \\ \hline
    \textbf{Evidence Combination 1} & \textbf{Evidence Combination 2} \\
    \texttt{\lbrack wiki/Lily\_James:1\rbrack} She studied acting at the Guildhall School of Music and Drama in London and began her acting career in the British television series Just William (2010) & \texttt{\lbrack wiki/Lily\_James:0\rbrack} Following her supporting role as Lady Rose MacClare in the period drama series Downton Abbey (2012–2015), James had her film breakthrough playing the titular role in the fantasy film Cinderella (2015). \\
    & \texttt{\lbrack wiki/Downton\_Abbey:0\rbrack} Downton Abbey is a British historical period drama television series set in the early 20th century, created by Julian Fellowes.  \\ \hline
    \end{tabular}
\egroup
    \centering
    \caption{Example FEVER dataset instance comprising a claim and multiple correct evidence sentence combinations}
    \label{fig:fever_example}
\end{figure}

\subsection{Task Definition}
We start by formally defining the FEVER task for which we will be generating adversarial instances for. A prediction for a FEVER instance (example in Figure~\ref{fig:fever_example}) that comprises a claim sentence $x$, 
correct (gold) label $y$, predicted label $\hat{y}$, 
set of correct evidence sentence combinations 
$\mathbf{E} = \lbrack E_1,\ldots, E_k \rbrack$ and predicted evidence sentence combination $\hat{E}$, it is scored as using equations~\ref{eqn:fever_icorrect} and \ref{eqn:fever_evcorrect}. 
\begin{equation}
\label{eqn:fever_icorrect}
\text{Instance\_Correct}(y,\hat{y},\mathbf{E},\hat{E}) \overset{\underset{\mathrm{def}}{}}{=} y=\hat{y} \wedge \big(y=\textsc{NotEnoughInfo} \vee \text{Evidence\_Correct}(\mathbf{E},\hat{E})\big)
\end{equation}

For claims labeled \textsc{NotEnoughInfo}, no evidence is required. Some claims require the combination of multiple sentences to be marked as fully supported or refuted: all sentences must be present in order for the submission to be marked as correct. As there may be multiple correct combinations of evidence sentences that can be used to \textsc{Support} or \textsc{Refute} a claim, 
at least one of these combinations 
needs to be predicted in its entirety 
for the prediction to be considered correct. \footnote{The full FEVER scorer is available as a stand-alone package on GitHub: \url{https://github.com/sheffieldnlp/fever-scorer} and can be installed with \texttt{pip install fever-scorer}}

\begin{equation}
    \label{eqn:fever_evcorrect}
    \text{Evidence\_Correct}(\mathbf{E}, \hat{E}) \overset{\underset{\mathrm{def}}{}}{=} \bigvee_{E\in\mathbf{E}} \big( \bigwedge_{e\in E} e \in \hat{E}\big)
\end{equation}

Over a set of predictions made on labeled instances, $\mathcal{Y}$, the FEVER score is defined as:

\begin{equation}
    \text{FEVER}(\mathcal{Y}) \overset{\underset{\mathrm{def}}{}}{=} \frac{1}{| \mathcal{Y} |}\sum_{(y, \hat{y}, \mathbf{E}, \hat{E}) \in \mathcal{Y}} \mathbb{I}\lbrack \text{Instance\_Correct}(y, \hat{y}, \mathbf{E}, \hat{E})\rbrack   
\end{equation}

For diagnostic purposes, we also report label accuracy (ignoring the need for correct evidence):

\begin{equation}
    \text{Accuracy}(\mathcal{Y}) \overset{\underset{\mathrm{def}}{}}{=} \frac{1}{| \mathcal{Y} |}\sum_{(y, \hat{y}, \mathbf{E}, \hat{E}) \in \mathcal{Y}} \mathbb{I}\lbrack y = \hat{y} \rbrack   
\end{equation}

The FEVER2.0 shared task requires \emph{breakers} (participants generating adversarial instances) $b \in B$, to generate and submit a balanced set (i.e.\ same number of instances per class) of novel claims $\tilde{{X}}_b = \{x_{b,i}\}^N_{i=1}$, a label and set of 
evidence sentence combinations $\tilde{{Y}}_b = \{(y_{b,i}, \mathbf{E}_{b,i})\}^N_{i=1}$. The number of novel claims in the submission will be limited: this will allow for manual evaluation to ensure that each submitted adversarial instance meets the shared task guidelines\footnote{An up-to-date version of the guidelines are available on the FEVER website: \url{http://fever.ai/task.html}}.  The adversarial 
instances generated that meet the guidelines 
(hereafter referred to as accepted instances: $X_b$, $Y_b$)
are then submitted to all systems from the builders, and the predictions made by these systems are used for scoring. 

We evaluate the adversarial instances from each breaker (ignoring the ones that failed to meet the guidelines) in two ways.
The \emph{potency} of the adversarial instances submitted by a breaker is defined as the average error rate over all the of predictions made by all systems $s\in{S}$ for labeled accepted instances generated by a breaker $b$, $\mathcal{Y}_{s,b}$:

\begin{equation}
\text{Potency}(b) \overset{\underset{\mathrm{def}}{}}{=} \frac{1}{|S|}\sum_{s \in S}\big( 1 - \text{FEVER}(\mathcal{Y}_{s,b}) \big)
\end{equation}

To penalize breakers for submitting claims that do not meet the guidelines (such as those which are ungrammatical or incorrectly labeled), the potency score will be scaled by the acceptance rate ($r_{accept}=\frac{|X_b|}{|\tilde{X}_b|}$) of the claims submitted: 

\begin{equation}
\text{Adjusted\_Potency}(b) \overset{\underset{\mathrm{def}}{}}{=} r_{accept} \times \text{Potency}(b)
\end{equation}

The \emph{resilience} of a system $s$ is defined as the \textsc{FEVER} score over all the accepted instances generated by all the breakers:

\begin{equation}
\text{Resilience}(s) \overset{\underset{\mathrm{def}}{}}{=} \text{FEVER}\big(\bigcup_{b \in B} \mathcal{Y}_{s,b}\big)
\end{equation}

\subsection{Method}
\label{sec:rules}
The use of human annotators to construct large scale datasets for natural language processing risks inducing biases and artifacts as the workers adopt strategies and shortcuts to generate examples. We hypothesize that the annotators for FEVER used a common set of constructions that could be exploited to generate a new set of adversarial claims. Counting the bigrams within the claims in training set, we identify the following forms: "is a", "an American", "was born", "directed by", "starred in" are some of the most frequently occurring.

Using these common claim patterns, we apply a collection of rule based transformations to the claims to generate new instances that do not reflect the patterns available in the training set. We evaluate three different groups of transformations: one where entailment between the claim and evidence is preserved, and two where it is reversed through negation (see examples in Table~\ref{tab:example}). The first set of transformations are entailment-preserving rewrites of the claim. There are similar to the methods that the annotators would have used when generating the FEVER dataset and includes simple techniques such as switching from active to passive voice, and adding pronouns. The second set of transformations were simple negations to the claim, performed mostly through a simple negation the sentences' verb phrase and reversing any \textsc{Supported} label to \textsc{Refuted} and vice versa. For some claim types, intensifiers such as `certainly' and `definitely' were also included in the sentence where this would not affect the entailment relation (as defined in Appendix~A of \cite{Thorne2018a}). The third set of rules combine both the entailment-altering rewrites and the simple negations (without the inclusion of any intensifiers), and we refer to them as complex negations. 

We constructed 65 rules that matched the most common claim patterns, for example: $X$ is a $Y$, $X$ was an $Y$, $X$ was a $Y$, $X$ was an $Y$, $X$ was directed by $Y$, $X$ died on $Y$, $X$ died in $Y$, $X$ was born in $Y$, $X$ is an American $Y$. 23 of these rules were entailment preserving, 19 were simple changes to negate claims, 23 were complex changes that also negated the claims. We release all of the rules that were used in this paper on our GitHub repository.

\begin{table}[]
\centering
\begin{tabular}{@{}llp{5.5cm}@{}}
\toprule
\multicolumn{1}{c}{\textbf{Transformation}} & \multicolumn{1}{c}{\textbf{Pattern}}  & \multicolumn{1}{c}{\textbf{Template}} \\ \midrule
Entailment Preserving & (.+)~\text{is~a}~(.+) & There exists a \$2 called \$1 \\
                      & (.+)~(?:\text{was~|~is})?~\text{directed by}~(.+) & \$2 is the director of \$1  \\ \midrule

Simple Negation & (.+)~was~an~(.+)& \$1 was not an \$2 \\
 & (.+)~\text{was born in}~(.+)& \$1 was never born \\ \midrule
Complex Negation & (.+)~(?:was~|~is)?~directed by (.+) & There is a movie called \$1 which wasn't directed by \$2 \\ 
 & (.+)~an~American~(.+) & \$1 \$2 that originated from outside the United States. \\
  \bottomrule \\
\end{tabular}
\caption{Example rule-based attacks that preserve the entailment relation of the original claim (within the definition of the FEVER shared task), perform simple negation and more complex negations. The output new claims that are sufficiently different to confuse the classifier. The matching groups within the regular expression are copied into the template (variables begin with \$).}
\label{tab:example}
\end{table}





\section{Experimental Setup}
We evaluate pretrained models for the FEVER shared task and their robustness to the adversarial instances we introduce in this paper. In this section, we compare the baseline model from \citet{Thorne2018a} (TF-IDF information retrieval with a decomposable attention model for natural language inference), a modified version of this model with an ESIM+ELMo for the NLI component, and the top four highest performing scoring system from the 2018 task \citep{nie2019combining, yoneda-EtAl:2018:FEVER, hanselowski-EtAl:2018:FEVER, malon:2018:FEVER}. For the baseline system, we adapted the code from \citep{Thorne2018a}, using the ESIM+ELMo implementation from AllenNLP \citet{Gardner2017} and trained both the Decomposable Attention and ESIM+ELMo models with the recommended parameters  for each model as included in the AllenNLP implementation. For the four highest scoring systems from the first shared task, we used the pretrained models released by each of the participants.

The adversarial claims are generated by applying rule-based transformations (described in Section~\ref{sec:rules}) to existing FEVER instances. 
In this paper, we made observations on the FEVER training set, counting common bigrams, that informed the creation of transformation rules that when applied to the dataset would create a large number of instances. 
In our experiments, we chose to retain the original evidence and only make changes to the claim text because it mitigates the need to expend effort annotating 
new evidence. 
We apply these rules to both the FEVER development and test splits of the dataset split presented in \citet{Thorne2018a} \footnote{Note: These are the development split and test split from the paper containing $9999$ examples each that were later concatenated, forming the development set of the first FEVER shared task} to generate a two datasets of novel claims. 
In Section~\ref{sec:results}, we evaluate the effects of the rule-base adversary using the claims from the development split and reporting the impact on both the NLI and Evidence Retrieval components. 
We report changes in performance over the entire pipeline as well as 
in an oracle environment where only the NLI component is tested.
For this oracle evaluation, we replace the evidence retrieval module 
with correct evidence as labeled by the annotators. 
For claims labeled as \textsc{NotEnoughInfo} in the oracle, we sample sentences from the nearest Wikipedia page to the claim (using TF-IDF, consistent with the \textsc{NearestP} approach reported by \citet{Thorne2018a}).  
We report the changes in FEVER Score over the end-to-end pipeline system, as well as report the changes in accuracy of the NLI model and the precision, recall and $F_1$ of the evidence retrieval component.
In Section~\ref{sec:sample}, we create a sample submission for the FEVER2.0 shared task and report the \emph{potency} of our technique and the \emph{resilience} of the systems we evaluate against.


\section{Results}
\label{sec:results}
65 rule-based transformations were executed to the data from the development set. There were $2708$ instances that matched the regular expression patterns defined in these rules, which were used to generate $29495$ novel adversarial instances. We report results using the instances which matched the rules (column labeled \emph{Before} in Table~\ref{tab:allresults}) that were used to generate the novel adversarial instances (column labeled \emph{After}). In the new set of claims that were generated, only $15\%$ were labeled as \textsc{NotEnoughInfo} (as the negating rules were only applied to claims which were \textsc{Supported} or \textsc{Refuted}) which would preclude comparison against existing results as the FEVER evaluation which assumes balanced class distribution. We balance our data by discarding instances at random from the majority classes. In a more comprehensive breakdown in Section~\ref{sec:effect}, we report results on all instances, broken down by the types of rules used in the adversarial attack. 

\begin{table}[h]
\centering
\begin{tabular}{lcccccccccc}
\toprule
\multicolumn{1}{c}{\multirow{2}{*}{\textbf{Model}}} &  
 \multicolumn{4}{c}{\textbf{Accuracy}}                                                                & \multicolumn{4}{c}{\textbf{FEVER Score}}                                                                      \\
 & \multicolumn{1}{c}{\textbf{Dev}}  
 & \multicolumn{1}{c}{\textbf{Before}} 
 & \multicolumn{1}{c}{\textbf{After}} 
 & \multicolumn{1}{c}{\textbf{Delta}} 
 & \multicolumn{1}{c}{\textbf{Dev}} 
 & \multicolumn{1}{c}{\textbf{Before}} 
 & \multicolumn{1}{c}{\textbf{After}} 
 & \multicolumn{1}{c}{\textbf{Delta}} \\ \midrule
Oracle + DA & \emph{82.83} & 87.94 & 68.06 & -19.88 & = & = & = & = \\
Oracle + ESIM & \emph{84.63} & 86.93 & 66.87 & -19.16 & = & = & = & = \\ \midrule
TFIDF + DA & \emph{51.94} & 50.50 & 39.18 & -11.32 & \emph{28.63} & 26.88 & 18.67 & -8.21 \\
TFIDF + ESIM & \emph{54.01} & 51.30 & 39.89 & -11.41 & \emph{33.72} & 33.17 & 21.62 & -11.55 \\ \midrule
NSMN (UNC) & \emph{69.73} & 71.23 & 49.18 & -22.05 & \emph{66.59} & 67.96 & 46.26 & -21.70 \\
UCL  & \emph{76.65} & 76.97 & 53.78 & -23.19 & \emph{69.42} & 71.06 & 47.73 & -23.33 \\
Athene & \emph{67.23} & 67.96 & 38.80 & -29.16 & \emph{62.39} & 63.99 & 34.13 & -29.86 \\ 
Papelo & \emph{75.60} & 74.75 & 55.73 & -19.02 & \emph{72.99} & 72.70 & 54.43 & -18.27 \\ \bottomrule \\
\end{tabular}
\caption{Summary of adversarial attack potency on three models using the FEVER development data split. For the Before and After columns, the figures are reported on the subset of the data used for the rule based transformation. We also report accuracy on the entire development split reported in \citet{Thorne2018a} ($9999$ instances in total, \textbf{Dev}). For the case of the oracle evidence retrieval component, FEVER Score is equal to accuracy. 
}
\label{tab:allresults}
\end{table}

We observe that in the oracle environment (using manually labeled evidence), both the ESIM and Decomposable Attention models for natural language inference suffer a stark decrease in accuracy when making predictions on the adversarial examples. The decrease is less pronounced when considering the full pipeline (i.e. with a non-perfect evidence retrieval component), which reduces the upper bound for the FEVER Score and accuracy. This is due to the noise introduced by the evidence retrieval component documented in Table~\ref{tab:evidencepr}.

\begin{table}[h]
\centering
\begin{tabular}{lccccccccc}
\toprule
\multicolumn{1}{c}{\multirow{2}{*}{\textbf{Model}}} &  
\multicolumn{3}{c}{\textbf{Precision}}    &
 \multicolumn{3}{c}{\textbf{Recall}}      & 
 \multicolumn{3}{c}{\textbf{F1}}                                                                      \\
\multicolumn{1}{c}{}                                 & 
\multicolumn{1}{c}{\textbf{Before}} & \multicolumn{1}{c}{\textbf{After}} & \multicolumn{1}{c}{\textbf{Delta}} &
 \multicolumn{1}{c}{\textbf{Before}} & \multicolumn{1}{c}{\textbf{After}} & \multicolumn{1}{c}{\textbf{Delta}} & \multicolumn{1}{c}{\textbf{Before}} & \multicolumn{1}{c}{\textbf{After}} & \multicolumn{1}{c}{\textbf{Delta}} \\ \midrule
TFIDF & 9.37& 8.62 & -0.75 & 39.07 & 35.30 & -3.77 & 15.12 & 13.86 & -1.26  \\
UNC & 42.49  & 45.38 & +2.89 & 75.63 & 69.74 & -5.89 & 54.41 & 54.98 & +0.57  \\
UCL  & 38.45  & 31.50 & -6.95 & 84.86 & 79.62 & -5.24  & 52.93 & 45.14 & -7.80  \\
Athene  & 26.50  & 25.53 & -0.97 & 88.94 & 86.66 & -2.58 & 40.84 & 39.44 & -1.40 \\
Papelo & 95.21 & 96.64 & +1.43 & 68.72 & 48.49 & -20.23 & 79.82 & 64.85 & -14.97 \\
\bottomrule \\
\end{tabular}
\caption{Effect of adversarial attacks on the evidence retrieval component of the pipelines considering sentence-level accuracy of the evidence.}
\label{tab:evidencepr}
\end{table}

The adversarial examples did not affect the information retrieval component of some systems to the same extent as the NLI component. Most systems incorporated either TF-IDF of keyword matching in their information retrieval component
and thus they are little affected by our
rule-based transformations which adding mostly stop words and reordering the words within the sentence. The only expecption is  Papelo, which even though uses TF-IDF to retrieve documents, its sentence-retrieval component made use of an entailment classifier. While this approach maintained very high precision with the newly generated instances, the recall decreased indicating a brittleness for instances that that differed from the distribution of the training set. 

\subsection{Effect of Transformation Types}
\label{sec:effect}
In Section~\ref{sec:rules}, we introduced three types of rule-based transformations (entailment-preserving paraphrasing, simple negations, and complex negations that combine 
the first two) that were used to generate the adversarial instances. Each of these rule-based transformations may expose different blind-spots and limitations within the models: we evaluate these in isolation and present a summary in Table~\ref{tab:breakdown} where we show that the systems under consideration exhibited a reduction in accuracy for each our rule-based transformation types.

\begin{table}[h]
\begin{tabular}{llcccccc}
\toprule
\multicolumn{1}{c}{\multirow{2}{*}{\textbf{Model}}} & \multicolumn{1}{c}{\multirow{2}{*}{\textbf{Transformation}}} & \multicolumn{3}{c}{\textbf{Accuracy}} & \multicolumn{3}{c}{\textbf{FEVER Score}} \\
\multicolumn{1}{c}{} & \multicolumn{1}{c}{} & \multicolumn{1}{c}{\textbf{Before}} & \multicolumn{1}{c}{\textbf{After}} & \multicolumn{1}{c}{\textbf{Delta}} & \multicolumn{1}{c}{\textbf{Before}} & \multicolumn{1}{c}{\textbf{After}} & 
\multicolumn{1}{c}{\textbf{Delta}} \\ \midrule
\multirow{3}{*}{Oracle+DA} & Entailment Preserving & 88.69 & 76.80 & -11.89  & =  & = & = \\
 & Simple Negations & 89.03 & 56.97 & -32.06 & = & = & =  \\
 & Complex Negations & 88.50 & 48.85 & -39.65 & = & = & = \\ \midrule
 
\multirow{3}{*}{Oracle+ESIM} & Entailment Preserving & 88.27 & 72.11 & -16.16 & = & = & =  \\
& Simple Negations & 88.65 & 57.93 & -30.72 & = & = & = \\ 
& Complex Negations & 86.49 & 53.91 & -32.58 & = & = & = \\ \midrule

\multirow{3}{*}{TFIDF+DA} & Entailment Preserving  & 52.42 & 43.89 & -8.53 & 27.63 & 20.11 & -7.52  \\
 & Simple Negations  & 68.18 & 53.58 & -14.60 & 31.90 & 25.34 & -6.56   \\
 & Complex Negations  & 66.96 & 47.70 & -19.56 & 30.46 & 20.29 & -10.17   \\ \midrule
 
\multirow{3}{*}{TFIDF+ESIM} & Entailment Preserving  & 52.50 & 43.52 &  -8.98 & 33.61 & 22.91 & -10.70  \\
& Simple Negations  & 61.29 & 48.70 & -12.59 & 33.29 & 25.24 & -8.05 \\ 
& Complex Negations  & 59.56 & 47.77 & -11.79 & 31.76 & 21.15 & -10.61 \\ \midrule

\multirow{3}{*}{UNC} & Entailment Preserving  & 71.91 & 51.19 & -20.72 & 68.69 & 48.21 & -20.48   \\
  & Simple Negations   & 75.54 & 53.72 & -21.82 & 70.75 & 49.16 & -21.59  \\
  & Complex Negations  & 75.24 & 44.49 & -30.75 & 70.56 & 40.67 & -29.89  \\ \midrule
  
\multirow{3}{*}{UCL} & Entailment Preserving   & 78.24 & 56.75 & -21.49 & 72.32 & 50.25 & -22.07   \\
  & Simple Negations    & 82.17 & 56.63 & -25.54 & 73.31 & 47.84 & -25.47  \\
  & Complex Negations   & 82.36 & 49.96 & -32.40 & 73.84 & 41.86 & -31.98  \\ \midrule
  
\multirow{3}{*}{Athene} & Entailment Preserving  & 68.60 & 40.32 & -28.28 & 64.85 & 35.40 & -30.45  \\
  & Simple Negations   & 71.52 & 56.95 & -14.57 & 65.62 & 50.58 & -15.04   \\
  & Complex Negations   & 71.21 & 53.00 & -18.21 & 65.70 & 46.45 & -19.25  \\ \midrule
  
\multirow{3}{*}{Papelo} & Entailment Preserving  & 74.00 & 58.61 & -15.39 & 71.99 & 57.21 & -14.78 \\
  & Simple Negations   & 68.54 & 42.87 & -25.67 & 65.40 & 40.37 & -25.03   \\
  & Complex Negations   & 69.27 & 30.02 & -39.26 & 66.39 & 28.81 & -37.58  \\ 
  \bottomrule \\
\end{tabular}
\caption{Breakdown of the potency of the three classes of transformation for each model. For the models with oracle information retrieval, the FEVER Score is equal to the label accuracy.}
\label{tab:breakdown}
\end{table}

We observe in all oracle and pipeline cases (with the exception of the Athene system) that the reduction in accuracy for the claims generated through label-preserving transformations was lower than the claims generated by the label-altering transformations. The effect of this is also exacerbated by the fact that the negated claims is only a binary classification rather than three-way for the entailment preserving claims. This may be revealing an inherent bias in the models similar to the one discussed by \citet{Naik2018}, where models perform poorly for antonymous examples due to a dependence on word-overlap as a feature, which we also observe with our negations. We also observe that 3 of top 4 ranking systems from the shared task (UNC, UCL and Papelo) exhibit a reduction in FEVER scorer for the negated instances. For the baseline (TFIDF+DA/ESIM) models, simple negations had a lower impact on the FEVER score than entailment preserving claims, possibly due to the noise coming from the poor evidence retrieval method (under oracle evaluation, this was not observed). 

The Athene system had a high error rate for the entailment preserving transformations: even though the model had an evidence recall of $86.37\%$ for the adversarial instances, the FEVER score reduced by $30.45\%$. On inspection, this model is mostly predicting \textsc{Supported} for our adversarial instances. In contrast, the Papelo system had the lowest reduction ($15.39\%$) in FEVER score for entailment preserving adversarial instances, despite a reduction in evidence recall of $23.17\%$. This model also exhibited a similar behaviour to the Athene model: while a large number of supported claims were correctly classified, this model predominantly predicted \textsc{NotEnoughInfo} for the adversarial instances. As this class does not require evidence to be correctly scored, `falling back' to it resulted in a higher score for balanced data. This also explains the low FEVER score for complex negations (a reduction of $37.58\%$), in this evaluating this class of transformation, we only considered \textsc{Supported} and \textsc{Refuted} claims while the model was mostly predicting \textsc{NotEnoughInfo}.




\section{Sample FEVER2.0 Shared Task Submission}
\label{sec:sample}
We construct a balanced sample submission for the break-it phase of the FEVER2.0 shared task. We applied our rule-based transformations to the test set from \cite{Thorne2018a} and then perform a stratified sample yielding 1000 instances. We compare this to a baseline that comprises a sample of the unmodified claims that reflects the current error rate of the systems prior to our adversarial attacks.

To verify the suitability of our method, we performed a manual evaluation of the claims generated by the rule based transformation. $30\%$ of the instances submitted was inspected and labelled as to whether the claims were grammatical. We found that $270$ (90\%) of the claims met the submission guidelines and $30$ ($10\%$) of the claims did not. Of the claims that did not meet the guidelines due to being nonsensical or non-grammatical: $1$ was generated from an original claim that had a grammatical error; $8$ were made from claims with more complex language that was not accounted for when designing the rule-based transformations; and $21$ had a systematic error where the third-person neutral pronoun \emph{it} was used for claims about people. 

\begin{table}[t]
\centering
\begin{tabular}{@{}llccc@{}}
\toprule
\textbf{Rank} & \textbf{Method} & \textbf{\begin{tabular}[c]{@{}l@{}}Potency\\ (\%)\end{tabular}} & \textbf{\begin{tabular}[c]{@{}l@{}}Accept\\ Rate (\%)\end{tabular}} & \textbf{\begin{tabular}[c]{@{}l@{}}Adjusted \\ Potency (\%)\end{tabular}} \\ \midrule
1 & Our Method (Rule-based Transformation) & 62.58 & 90 & 56.32 \\
2 & Baseline (Unmodified Instances) & 45.48 & 97 & 44.12 \\ \bottomrule \\
\end{tabular}
\caption{Sample leader-board for the FEVER2.0 breakers. The adjusted potency score is an estimate based on the results of the manual inspection of the generated claims.}
\end{table}

\begin{table}[t]
\centering
\begin{tabular}{@{}llc@{}}
\toprule
\textbf{Rank} & \textbf{System} & \textbf{Resilience (\%)} \\ \midrule
1 & Papelo & 63.16 \\
2 & UCL & 58.31 \\
3 & UNC & 57.16 \\
4 & Athene & 47.00 \\
5 & TFIDF+ESIM & 26.83 \\
6 & TFIDF+DA & 23.37 \\ 
\bottomrule \\
\end{tabular}
\caption{Sample leader-board for the FEVER2.0 systems ranked by the resilience (average FEVER score over all breaker submissions). }
\end{table}

Considering the performance of the systems we test, we also present a leader-board ranked by the resilience (average FEVER score for all breaker methods). While the Papelo system was ranked $4^{\text{th}}$ on the FEVER shared task, this model performs the best 
in this evaluation. In Table~\ref{tab:breakdown}, we presented a breakdown of results which indicated, that under entailment preserving transformations, this model performed better for the other systems we tested against. Entailment-preserving transformations accounted for around $60\%$ of the claims sampled our submission\footnote{Only the entailment preserving rules generated claims of all three classes \text{Supported}, \textsc{Refuted} and \textsc{NotEnoughInfo} whereas the negating transformations were only applied to the \textsc{Supported} and \textsc{Refuted} claims} which may account for this result. The UCL system (ranked $2^\text{nd}$ in the shared task) also performs better than UNC system (ranked $1^\text{st}$ in the shared task) in both this sample evaluation and the breakdown in Table~\ref{tab:breakdown}. This may arise from the fact that both teams make use of the development data that we used to generate our claims differently (i.e. the pretrained UCL model we used incorporated development data to increase the size of the training set whereas other models did not). As it is not possible to control how this data is used in the FEVER2.0 shared task, our method for generating claims from transforming existing instances may perform better (i.e. have more potency) when applied to instances that have not previously been released to the public.

\section{Conclusions}
In this report, we explored the behaviour of systems for Fact Extraction and VERification under adversarial attacks by applying simple rule-based perturbations to existing instances.
We showed that adversarial examples can be generated through very simple transformations that are low cost and generate 
grammatically correct instances. We hope that the findings that we present in this paper help continue the discussion on improving models to be robust and resilient to data from outside of the training distribution. The FEVER2.0 shared task is just one forum for discussing the robustness of models against adversarial examples. We hope that the techniques in this paper help inspire participation in the task and that the data generated by the shared task may be used by researchers in other disciplines to help characterize the data used to train their models and the potential risks of attack that these models may face when deployed in the wild.

\section*{Acknowledgements}
The authors wish to thank Trevor Cohn, Tim Baldwin and the FEVER Organizers for their helpful advice.

\bibliographystyle{plainnat}
\bibliography{mendeley}

\end{document}